# <CT>It's distributions all the way down!

<CA>Mark T. Keane[a] and Aaron Gerow[b]

<CAA>[a]School of Computer Science & Informatics, University College Dublin; Belfield, Dublin 4, Ireland; [b]School of Computer Science & Statistics, University of Dublin, Trinity College, College Green, Dublin 2, Ireland.

mark.keane@ucd.ie
gerowa@tcd.ie

www.csi.ucd.ie/users/mark-keane
www.scss.tcd.ie/~gerowa

<C-AB>**Abstract:** The textual, big-data literature misses Bentley, O'Brien, & Brock's (Bentley et al.'s) message on distributions; it largely examines the first-order effects of how a single, signature distribution can predict population behaviour, neglecting second-order effects involving distributional shifts, either between signature distributions or within a given signature distribution. Indeed, Bentley et al. themselves under-emphasise the potential richness of the latter, within-distribution effects.

<C-Text Begins>

It has been reported, possibly as aprocrypha, that when South Sea islanders were asked to explain their cosmology about what supported the World Turtle (who supported the world), one bright spark replied: "It's turtles all the way down!" The take-home message from the target article is analogously: "It's distributions all the way down!" Though the message may seem obvious, it is interesting to note that most of the current textual, big-data literature either doesn't get it or fails to appreciate its richness.

The lion's share of the textual, big-data literature (that analyzes, e.g., words, sentiment terms, tweet tags, phrases, articles, blogs) really concentrates on what could be called first-order effects rather than on second-order effects. *First-order effects* are changes in a single, statistical distribution of some measured text-unit (word, article, sentiment term, tag) – usually from some large, unstructured, online dataset – that can act as a predictive proxy for some population-level behaviour. So, for instance, Michel et al. (2011) showed how relative word frequencies in Google Books can reflect cultural trends; others have shown that distributions of Google-search terms can predict the spread of flu (Ginsberg et al. 2008) and motor sales (Choi & Varian 2012), and both numbers of news articles and tweet rates can be used to predict movie box-office revenues (Asur & Huberman 2010; Zhan & Skiena 2009). In these cases, the actual distribution of the text data is directly reflected in some population behaviour or decision.

*Second-order distributional effects* are changes across several, separate distributions that can predict population-level behaviour: usually, where these separate distributions span different time periods. Big-data research on such effects is a lot less common; to put it succinctly, most of the work focusses on either the wisdom of crowds (Surowiecki 2005) or herd-like behaviour (Huang & Chen 2006) but not on how a wise crowd becomes a stupid herd (for rare exceptions, see Bentley & Ormerod 2010; Bentley et al. 2012; Onnela & Reed-Tsochas 2010; Salganik et al. 2006). Bentley et al. recognise the importance of such second-order effects in stressing that population decision-making may shift from one signature distribution to another. However, we believe that there is more to be said about such second-order effects.

For instance, second-order effects can be further divided into between-distribution and within-distribution effects. *Between-distribution effects* concern the types of changes, mentioned by Bentley et al., where there is change from one signature distribution to another over time (movement between the quadrants of their map): for example, from the Gaussian distribution of the wisdom of crowds to the power-law distribution of herd behaviour. In physical phenomena, such *phase transitions* are well-known as water turns to ice or as a heated iron bar becomes magnetised (see Barabasi & Gargos 2002). Bentley et al. raise the prospect that there are a host of similar phase transitions in human population behaviour, transitions that can be tracked and predicted by mining various kinds of big data. We find this prospect very exciting given the notable gap in the big-data literature on such effects.

Bentley et al. say a lot less about *within-distribution effects*, the systematic changes that may occur in the properties of a given signature distribution from one time period to the next; for example, the specific properties of separate power-law distributions of some measured item might change systematically over time. Research on such within-distribution effects in textual, big-data research is even rarer than that on between-distribution effects – one of these rarities being our own work on stock-market trends (Gerow & Keane 2011). Gerow & Keane (2011) found systematic changes in the power-law distributions of the language in finance articles (from the BBC, *New York Times,* and *Financial Times*) during the emergence of the 2007 stock-market bubble and crash, based on a corpus analysis of 18,000 online news articles over a 4-year period (10M+ words). Specifically, they found that week-to-week changes in the power-law distributions of verb-phrases correlated strongly with market movements of the major indices; for example, the Dow Jones Index correlated ($r = 0.79$) with changes in the 8-week-windowed, geometric mean of the alpha terms of the power laws. These within-distribution changes showed that, week-on-week, journalists were using a progressively narrower set of words to describe the market, reporting on the same small set of companies using the same overwhelmingly positive language. So, as the agreement between journalists increased, the power-law distributions changed systematically in ways that tracked population decisions to buy stocks.

In Bentley et al.'s map, this analysis is about tracking movement *within* the southeast quadrant. Naturally, we also find the prospect of such within-distribution effects very exciting, especially given the lack of attention given to them thus far.

So, it really is "Distributions all the way down!" Current research is only scraping the surface of the possible distributional effects that may be mined from various online sources. Bentley et al. have provided a framework for thinking about such second-order effects, especially of the between-distribution kind, but there is also a whole slew of within-distribution phenomena yet to be explored.

<C-Text ends>

<Ref ends>